\documentclass{article}
\usepackage{spconf,amsmath,graphicx,hyperref}
\usepackage{amssymb,booktabs,multirow,enumitem}
\usepackage{graphicx}
\usepackage{subcaption}

\title{Once is Enough: Lightweight DiT-based Video Virtual Try-on via One-time Garment Appearance Injection}
\name{Yanjie Pan$^{1}$
  ~~ Qingdong He$^{2}$ 
  ~~ Lidong Wang$^1$
  ~~ Bo Peng$^3$
  ~~ Mingmin Chi$^1$$^\dagger$
  \thanks{$^\dagger$Corresponding author (mmchi@fudan.edu.cn)}}
\address{
\normalsize 
   $^1$School of computer science, Shanghai key laboratory of data science, Fudan University, China \\ \normalsize 
   ~~ $^2$Tencent Youtu Lab, China 
   ~~ $^3$Shanghai Ocean University, China 
}
\begin{document}
\ninept
\maketitle
\begin{abstract}
Video virtual try-on aims to replace the clothing of a person in a video with a target garment. Current dual-branch architectures have achieved significant success in diffusion models based on the U-Net. However, adapting them to diffusion models built upon the Diffusion Transformer remains challenging.
Initially, introducing latent space features from the garment reference branch requires adding or modifying the backbone network, leading to a large number of trainable parameters. Subsequently, the latent space features of garments lack inherent temporal characteristics and thus require additional learning. 
To address the above challenges, we propose a novel approach, OIE(Once is Enough), a virtual try-on strategy based on first-frame clothing replacement. Specifically, we employ an image-based clothing transfer model to replace the clothing in the initial frame. Subsequently, under the content control of the edited first frame, we utilize pose and mask information to guide the temporal prior of the video generation model in synthesizing the remaining frames sequentially.
Experiments show that our method achieves superior parameter efficiency and computational efficiency, while still maintaining leading performance under these constraints.
\end{abstract}
\section{Introduction}
\label{sec:intro}

Virtual try-on aims to replace the clothing worn by the main character in a video with target garments, while maintaining consistent motion behavior and visual appearance throughout the dynamic process and achieving complex interactions with human movements. Virtual try-on addresses the issue of trying on clothes for consumers on e-commerce platforms, effectively reducing transaction costs for both merchants and customers.

The video virtual try-on task fundamentally differs from its image-based counterpart in that it requires both cross-frame visual consistency of the garment appearance and semantic alignment with dynamic human poses and motions.

Previous virtual try-on methods~\cite{fang2024vivid,li2025magictryon,chong2025catv2ton,jiang2022clothformer,deng2023mv,dong2019fw,zheng2024dynamic} have relied on lightweight video or image generation models~\cite{rombach2022high,blattmann2023stable,goodfellow2020generative} as their backbone architecture, leading to the development of a mature and effective technical framework — the \textit{dual-branch architecture}. This architecture introduces a garment reference branch to extract appearance features of the clothing and integrates them into the main generation branch. By training feature interaction modules between the two branches, these methods~\cite{fang2024vivid,nguyen2025swifttry,he2024wildvidfit,xu2024tunnel} capture the temporal appearance and behavioral characteristics of the garments. Although such approaches have achieved significant results, they are limited by the representational capacity of the U-Net based backbone network, especially when it comes to rendering complex textures and details in human figures and garments.

Thanks to the strong capability of Diffusion Transformer(DiT)~\cite{peebles2023scalable,ho2020denoising} based video models~\cite{wan2025wan,kong2024hunyuanvideo} in capturing content, spatial, and temporal information, virtual try-on based on such architectures holds great promise. Some efforts have attempted to apply DiT-based generative models to the dual-branch framework and have achieved certain success.
However, they still face the challenge of excessive parameter count. The diffusion model based on the DiT architecture, used as the backbone network, leads to a significantly large number of parameters. Combined with the introduction of garment features through the reference branch and the need to train feature interaction modules, this makes the issue of model parameter scale even more critical.

    

To address the above issues, we propose a novel technical architecture — a \textit{first-frame guided single-branch framework}. Specifically, we utilize a powerful image-based garment editing model~\cite{jiang2024fitdit} to edit the first frame, and then propagate this edit to subsequent frames using a diffusion model fine-tuned with low-rank adaptation(LoRA)~\cite{hu2022lora}. This architecture decouples the problem of garment editing from that of preserving temporal consistency in video content, addressing each in the image editing stage and the video generation stage, respectively. Moreover, thanks to the separation of the garment reference branch and the use of LoRA fine-tuning techniques, we significantly reduce the number of parameters required in the video generation model, greatly lowering memory and computational resource consumption.

Our main contributions can be summarized as follows:
\begin{itemize}
    \item We propose a novel architecture to address the adaptability, training efficiency, and resource consumption challenges of DiT in video virtual try-on tasks.
    \item OIE strikes an optimal balance between performance and efficiency, achieving leading results in video virtual try-on with minimal computational overhead over the base model.
    \item We conduct comprehensive qualitative and quantitative experiments to evaluate the effectiveness and performance of our proposed method.
\end{itemize}

\section{Method}
\label{sec:method}

\begin{figure*}[ht]
\centering
\includegraphics[width=0.95\linewidth]{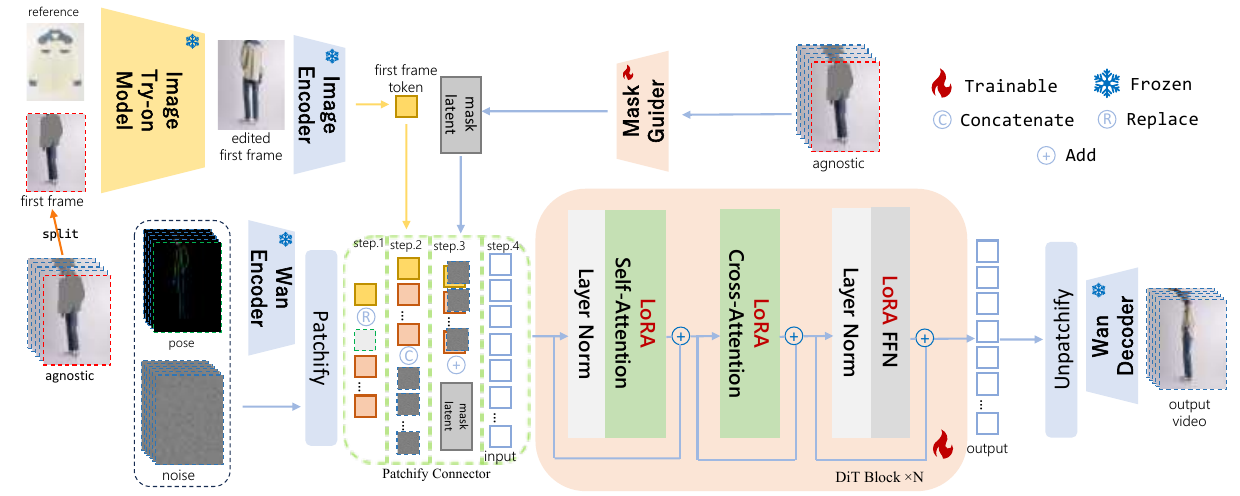}
\vspace{-5pt}
\caption{\textbf{Overall framework of OIE.} OIE leverages a pre-trained image virtual try-on model to inject garment appearance, and further adapts the input configuration on a LoRA-finetuned model to achieve a low-parameter, structure-preserving video virtual try-on strategy.}
\vspace{-15pt}
\label{fig:pipeline}
\end{figure*}

Our method aims to lightweight existing dual-branch DiT-based \textbf{OIE (Once is Enough)} frameworks by adopting a single-branch design. The overall framework of our proposed OIE is illustrated in Fig.\ref{fig:pipeline}. \textbf{Specifically, OIE~first edits the initial frame of the source video using a pretrained image virtual try-on model, and then leverages the first-frame diffusion capability of a large-scale video DiT to generate the full video try-on.} This design reduces the heavy resource demands of video DiTs while preserving garment fidelity and multi-scale consistency.

\subsection{Garment Appearance \textbf{One-time} Injection}
As we discussed in Sec.~\ref{sec:intro}, existing DiT-based frameworks adopt a dual-branch paradigm that repeatedly injects garment appearance features from the reference branch into the DiT backbone, resulting in substantial computational overhead. In addition, these methods require architectural adaptations of the backbone network to achieve better feature alignment, which further increase resource demands and model complexity. Consequently, reducing appearance feature injection frequency and minimizing structural modifications have become increasingly urgent to balance model efficiency and generation performance. \textbf{To fulfill this goal, inspired by Wan’s~\cite{wan2025wan} highly harmonious performance in video content synthesis and considering that the noise input already encodes video structural information, we propose a novel garment appearance injection strategy based on first-frame editing, as illustrated in Fig.~\ref{fig:pipeline}.}

Specifically, given the input background video $ x_{\text{agn}} $, we extract its first frame $ i_0 $ for injecting garment appearance information. For $ i_0 $, which is an image virtual try-on task, existing image virtual try-on models have the capability to capture fine-grained garment texture details and handle person-clothing interactions effectively. Therefore, we employ a pretrained image virtual try-on model FiTDiT $ M_{\text{FiT}} $~\cite{jiang2024fitdit} to perform appearance injection on $ i_0 $, obtaining a high-quality image try-on result $ i_r $. The formal process is formulated as:
\begin{equation}
    i_r = M_{\text{FiT}}(i_0, c_i, r_{clo})
\end{equation}
where $ c_i $ refers to the given textual instruction that uniformly describes the clothing change process, and $ r_{\text{clo}} $ denotes the front-view high-resolution image of the garment, serving as the source of detailed texture information for the clothing.

In order to minimize disruption to Wan's temporal priors in video synthesis and reduce information perturbation, we adopt Wan's image encoder $ E_{\text{img}} $ to obtain the latent space feature $ L_g $ of $ i_r $. 
Due to the edited first frame contains complete human skeletal structure, and accurate pose control is crucial for ensuring motion consistency in video virtual try-on tasks, we further employ Wan's VAE $ E_{vae} $ to obtain the latent representation $ L_p $ of the pose video $ v_p $. 
Subsequently, $ L_g $ is embedded as the first token into $ L_v $, resulting in a unified latent sequence $ L $ that incorporates both temporal dynamics and garment appearance information. The mathematical formulation is as follows:
\begin{equation}
L = R\left(\text{E}_{vae}(v_p),\ E_{\text{img}}(i_r)\right)
\end{equation}
where $ R $ denotes the embedding operation, and $ L \in \mathbb{R}^{N \times d} $.

\begin{table}[h]
\vspace{-5pt}
\resizebox{\linewidth}{!}{
\begin{tabular}{c|c|c|c}
\toprule
\multicolumn{1}{c|}{\textbf{Method}} &\multicolumn{1}{c|}{\textbf{Framework}} & \multicolumn{1}{c|}{\textbf{Reference Injection}} & \multicolumn{1}{c}{\textbf{No Structural Modifications}}  \\ 
\midrule
ViViD & dual branch &  N & \texttimes \\
CatV$^2$TON & dual branch & N & \texttimes \\
MagicTryOn & dual branch & N & \texttimes \\
Ours & single branch & 1 & \checkmark \\
\bottomrule
\end{tabular}}
\vspace{-10pt}
\caption{%
    \textbf{Method Complexity Analysis.} Analyze the complexity of method from two key perspectives: the number of garment appearance injection steps and whether the backbone network is modified during implementation.
}
\label{tab:condition_types}
\vspace{-10pt}
\end{table}
As shown in Table~\ref{tab:condition_types}, the proposed architecture offers the following advantages: a): Previous dual-branch approaches require injecting garment appearance information at every time step, whereas the proposed single-branch architecture only needs a single injection at the input stage. b): The dual-branch architecture often requires modifications to the backbone network due to the need for aligning garment features with the latent feature space. In contrast, the proposed single-branch architecture effectively avoids such architectural changes.

Specifically, the garment appearance injection strategy in dual-branch architectures urgently requires an adaptation module to align the garment features with the latent features of the diffusion model. Moreover, to achieve more harmonious synthesis results and ensure cross-frame consistency of the garment features in the video, such methods often need to further modify the backbone architecture in order to project the garment features into the latent feature space of the backbone. We observe that the significant performance improvement of DiT over the UNet architecture allows DiT to leverage its powerful temporal priors to maintain both consistency and quality in video synthesis. Therefore, our strategy of injecting garment information only at the first frame avoids any modification to the backbone network architecture, while achieving a balanced trade-off between efficiency and temporal consistency.

\subsection{Preserving Background Integrity}
Since the pose video focuses on human motion estimation and eliminates scene information from clothing-agnostic regions, it inevitably loses the layout information of the background.
In addition, maintaining low backbone perturbation while aligning with the garment appearance injection strategy is essential for improving model efficiency. To this end, we propose a lightweight background encoder to enhance the background information from the agnostic video, while avoiding any adaptation or modification to the backbone network.

The encoder $ E_m $ consists of four 3D convolutional layers, with channel dimensions of 32, 96, 192, and 256, respectively.  
To facilitate smoother guidance of the mask latent during training for video synthesis, we zero-initialize a linear layer within the mask guider. Specifically, as illustrated in Figure~4, the clothing-agnostic video $ x_{\text{agn}} $ is fed into the mask guider, and its output $ L_m $ is subsequently added to the input features concatenated at the second stage of the backbone network. The corresponding mathematical formulation is given as:
\begin{equation}
L = L + E_m(x_{\text{agn}}).
\end{equation}

To further reduce the number of trainable parameters when fine-tuning the diffusion model based on the DiT architecture, we apply LoRA to the backbone attention modules in Wan, including both the self-attention and cross-attention modules. The formulation is as follows:
\begin{equation}
\begin{aligned}
Q &= W_{Q} + A_Q B_Q^\top, \\
K &= W_{K} + A_K B_K^\top, \\
V &= W_{V} + A_V B_V^\top,
\end{aligned}
\end{equation}
where $ W_{Q}, W_{K}, W_{V} $ denote the pre-trained weight matrices of the original projection layers, and $ A_Q, B_Q, A_K, B_K, A_V, B_V $ are low-rank trainable matrices with rank $ r \ll \min(d, k) $. This parameter-efficient adaptation enables dynamic modulation of attention mechanisms while preserving the original model capacity. We further apply LoRA to the linear transformations in the feed-forward network (FFN), where each weight matrix $ W_{F0} $ is similarly parameterized as $ W_F = W_{F} + A_F B_F^\top $. This extension allows for lightweight fine-tuning across both attention and MLP components of the diffusion Transformer.

\subsection{Training Objective}
By leveraging flow matching to maintain equivalence with the maximum likelihood objective, the model is trained to learn the true velocity. The overall training objective $ \mathcal{L} $ is formulated as follows:
\begin{equation}
\mathcal{L} = \mathbb{E}_{c_{\text{txt}}, t, x_1, x_0} \left[ \| u(x_t, L, c_{\text{txt}}, t; \theta) - v_t \|_2^2 \right]
\end{equation}
where $ u(x_t, L, c_{\text{txt}}, t; \theta) $ denotes the output velocity predicted by the model.

\section{Experiments}
\label{sec:experiments}
\subsection{Datasets and Metrics}
We selected two publicly available video virtual try-on datasets, VVT and ViViD. Following the evaluation protocols used in prior works~\cite{chong2025catv2ton,li2025magictryon}, we conduct experiments on the ViViD, under paired and unpaired settings. In the paired setting, the reference garment matches the one worn by the human model, allowing us to evaluate the model's ability to capture fine-grained garment texture details. In contrast, in the unpaired setting, the reference garment differs from the one worn, which assesses the model's generalization and garment adaptation capability.

For efficiency evaluation, considering that different methods are built upon varying baselines, we conduct comparisons relative to each respective baseline. Regarding computational efficiency, we measure both FLOPs and actual inference latency. For parameter count, we compare not only the incremental parameters added over the base model, but also the number of trainable parameters during training and inference.

To evaluate video generation quality, we employ three widely adopted metrics in the virtual try-on domain: SSIM, LPIPS, and VFID~\cite{zhang2018unreasonable,wang2004image,carreira2017quo,hara2018can}. Specifically, SSIM measures the structural similarity between the generated and reference videos. VFID evaluates both temporal consistency and overall video synthesis quality. LPIPS is used to measure perceptual differences between pairs of images.

\subsection{Implementation Details}
We adopt the pre-trained weights from Wan2.1-I2V-14B-720P as the base model. During the training phase, each training video sample consists of 81 frames, with a batch size set to 1. The training is conducted for 30000 iterations. We employ the AdamW optimizer with a fixed learning rate of 1e-4.
For inference, we utilize FiTDiT~\cite{jiang2024fitdit} as the virtual try-on model for images, adhering to the official default settings for image inference. 
For video inference, the number of inference steps is set to 25. Additionally, to ensure a fair comparison, all model variants involved in the ablation studies are evaluated under the same hyperparameter configurations during inference.
\subsection{Main Results}

\begin{table}[h]
\vspace{-13pt}
\caption{%
    \textbf{Computational and Parameter Overhead on ViViD~\cite{fang2024vivid} between dual-branch methods and OIE across base models.} (ffirst row per column: base model). Base model of CatV$^2$TON not fairly reproducible due to removal of text attention layers.
}
\vspace{-5pt}
\resizebox{\linewidth}{!}{
\begin{tabular}{c|c|c|c|c}
\toprule
\multicolumn{1}{c|}{{\textbf{Methods}}} & 
\multicolumn{1}{c|}{{\textbf{FLOPs(G)}}} & 
\multicolumn{1}{c|}{{\textbf{s/it}}}&
\multicolumn{1}{c|}{{\textbf{Inference Param(B)}}} &
\multicolumn{1}{c}{{\textbf{Training Param(B)}}} \\
\midrule
Wan2.1~\cite{wan2025wan} & 193899.00 & 12.38 &  14.28602 & -  \\
MagicTryOn~\cite{li2025magictryon} & 206934.73 &13.69& 16.44604  &  16.44604 \\
Ours & 193899.00 &12.41& 14.36269  &  0.07994 \\ 
\midrule
Stable Diffusion 1.5~\cite{rombach2022high} & 338.75 & 0.05 &  0.85940 & -  \\
ViViD~\cite{fang2024vivid} & 960.60 & 2.07 &  2.20933 & 2.20933  \\
\midrule
CatV$^2$TON~\cite{chong2025catv2ton} & 11672.00 & 1.04 &  1.33655 & 1.33655  \\
\bottomrule
\end{tabular}}
\label{tab:efficiency_comparison}
\vspace{-12pt}
\end{table}
Table~\ref{tab:efficiency_comparison} demonstrates the efficiency advantage of the first-frame editing strategy for video virtual try-on. Compared with the base model, OIE introduces only a 0.50\% additional parameter overhead, which is significantly lower than MagicTryOn's 15.11\% and ViViD's 157.10\%. 
Moreover, in terms of inference efficiency, OIE incurs no noticeable additional cost in either FLOPs or actual inference time relative to the base model, substantially outperforming other approaches.

\begin{figure}[ht]
\centering
\includegraphics[width=0.75\linewidth]{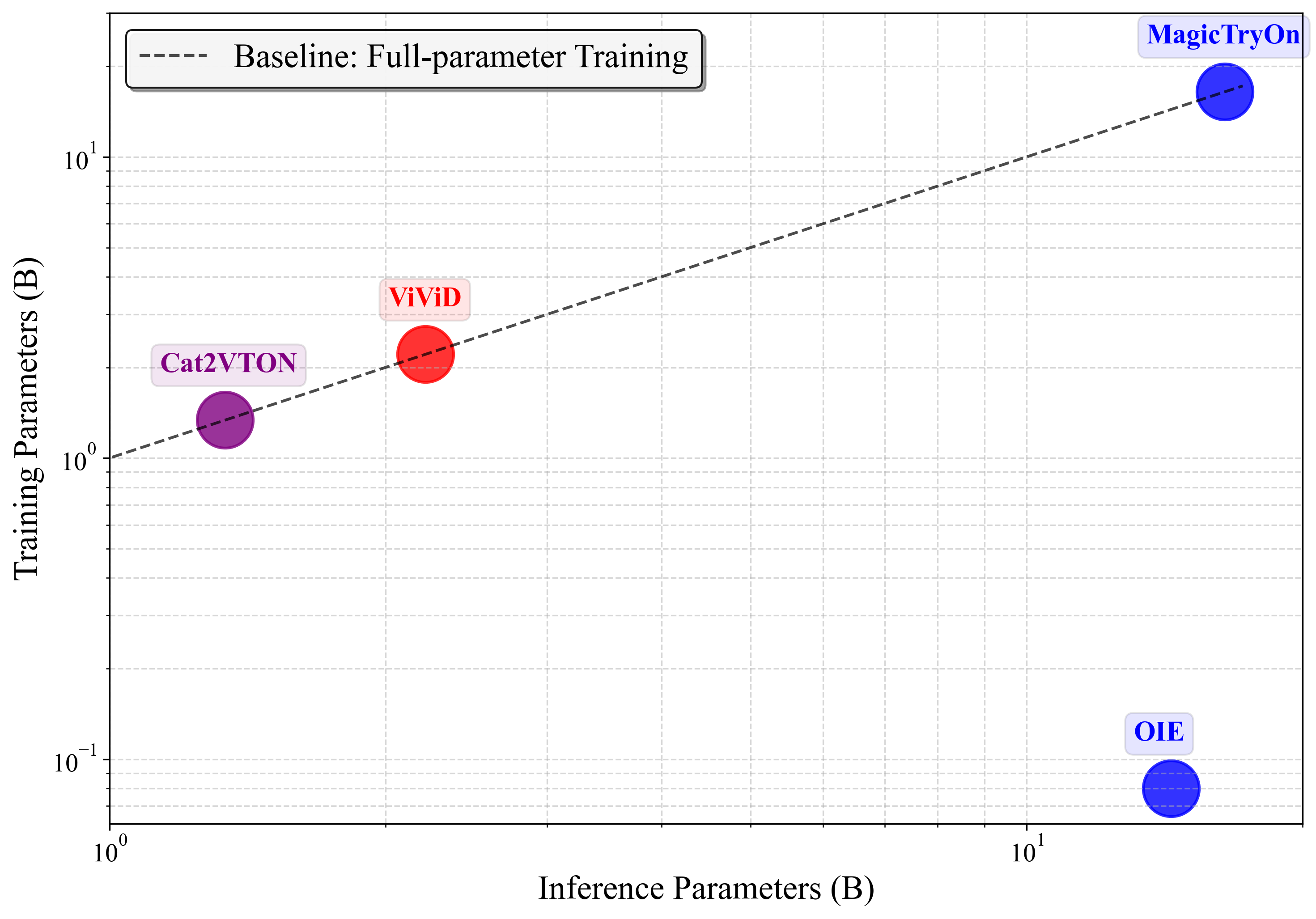}
\vspace{-5pt}
\caption{
\textbf{Comparison of Parameter Count.} Dual-branch methods exhibit a training-to-inference parameter ratio lying along the dashed diagonal line (slope = 1), indicating nearly identical parameter counts during training and inference.
}
\label{fig:param}
\end{figure}

\begin{figure}[ht]
\centering
\includegraphics[width=0.75\linewidth]{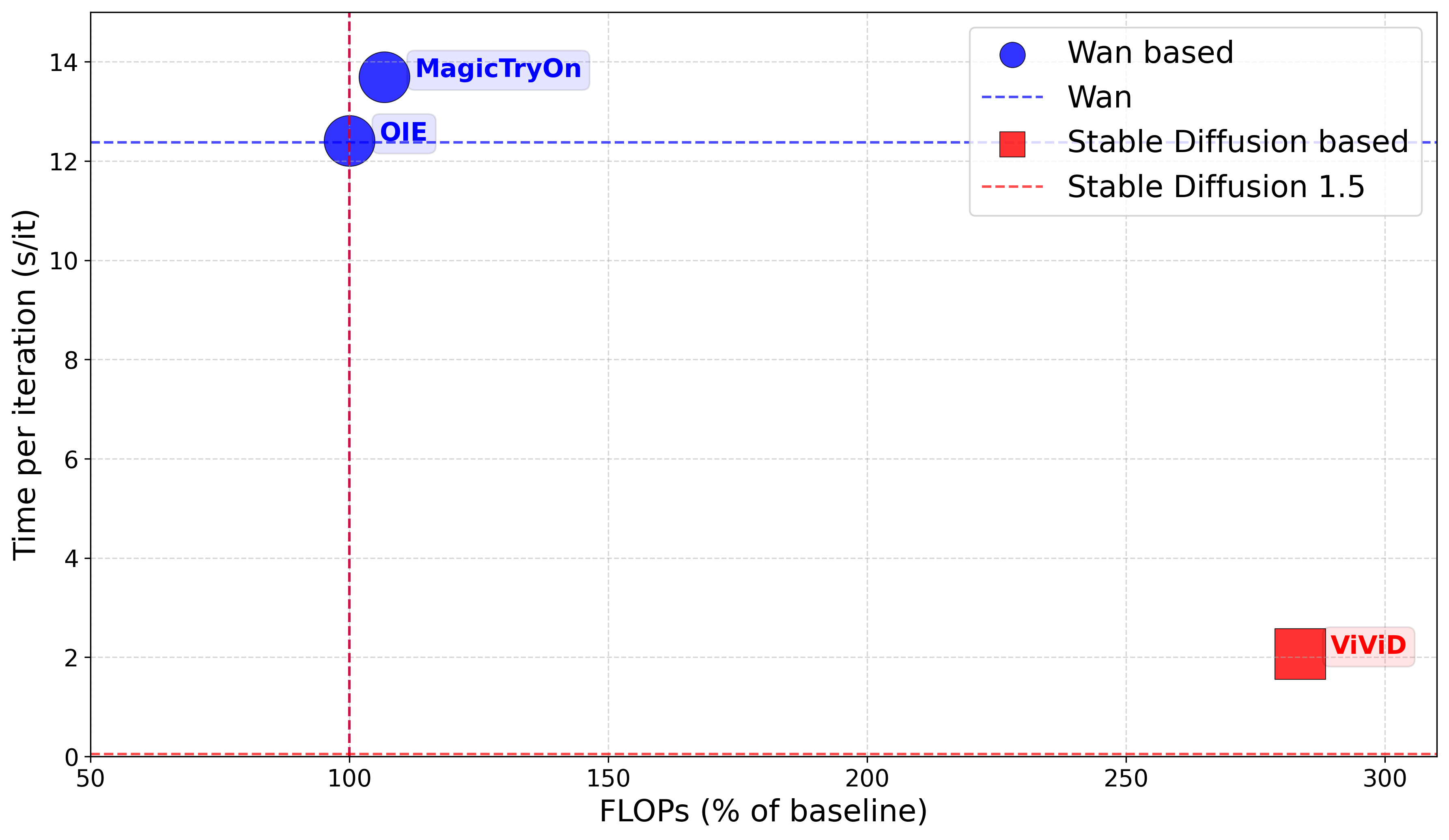}
\vspace{-5pt}
\caption{
\textbf{Comparison of Computational Burden.} Dashed lines denote baseline performance and dots represent models. OIE introduces negligible inference overhead compared to others. 
}
\vspace{-20pt}
\label{fig:compute}
\end{figure}

\begin{table}[h]
\vspace{-5pt}
\caption{%
    \textbf{Video Quality Comparison on ViViD.} The best and second-best results are marked with bold and underline, respectively. $p$ and $u$ denote the paired setting and unpaired setting.
}
\vspace{-5pt}
\resizebox{\linewidth}{!}{
\begin{tabular}{c|c|c|c|c}
\toprule
\multicolumn{1}{c|}{} & \multicolumn{3}{c|}{\textbf{paired}} & \multicolumn{1}{c}{\textbf{unpaired}}  \\ \cline{2-5}
\multicolumn{1}{c|}{\multirow{-2}{*}{\textbf{Methods}}} &  \multicolumn{1}{c|}{\textbf{VFID$^p$ ($\downarrow$)}}& \multicolumn{1}{c|}{\textbf{SSIM~($\uparrow$)}} & \multicolumn{1}{c|}{\textbf{LPIPS ($\downarrow$)}} &  \multicolumn{1}{c}{\textbf{VFID$^u$ ($\downarrow$)}} \\ 
\midrule
StableVITON~\cite{kim2024stableviton} & 34.2446 & 0.8019 &  0.1338 & 36.8985 \\
OOTDiffusion~\cite{xu2025ootdiffusion} & 29.5253 & 0.8087 &  0.1232 & 35.3170 \\
IDM-VTON~\cite{choi2024improving} & 20.0812 & 0.8227 & 0.1163 & 25.4972 \\ 
ViViD~\cite{fang2024vivid} &  17.2924 & 0.8029 & 0.1221 & 21.8032 \\
CatV$^2$TON~\cite{chong2025catv2ton} & 13.5962 & \underline{0.8727}  &  \textbf{0.0639} &19.5131 \\
MagicTryOn~\cite{li2025magictryon} & \underline{12.1988} & \textbf{0.8841} & 0.0815& \underline{17.5710} \\
Ours & \textbf{9.3983} &  0.8466 & \underline{0.0774} & \textbf{17.0831} \\
\bottomrule
\end{tabular}}
\vspace{-12pt}
\label{tab:comparision}
\end{table}

Figure~\ref{fig:param} illustrates the parameter comparison between training and inference stages. Existing dual-branch architectures (MagicTryOn, ViViD, CatV$^2$TON) require structural modifications to the backbone for garment feature integration and rely on the full DiT model, resulting in \textbf{identical parameter counts} during training and inference, implying limited potential for parameter-level efficiency gains during training. In contrast, OIE avoids direct injection of garment features into the backbone and instead employs LoRA fine-tuning, drastically \textbf{reducing the number} of trainable parameters. As demonstrated in Figure~\ref{fig:compute}, due to its minimal additional parameters, OIE \textbf{introduces negligible inference overhead}, with FLOPs and parameter usage increasing by less than 0.1\% compared to the base model. In contrast, dual-branch architectures introduce additional garment branches and feature fusion layers, which significantly increase computational cost and result in substantial inference overhead beyond the base model. 

\begin{figure}[ht]
\centering
\includegraphics[width=1.0\linewidth]{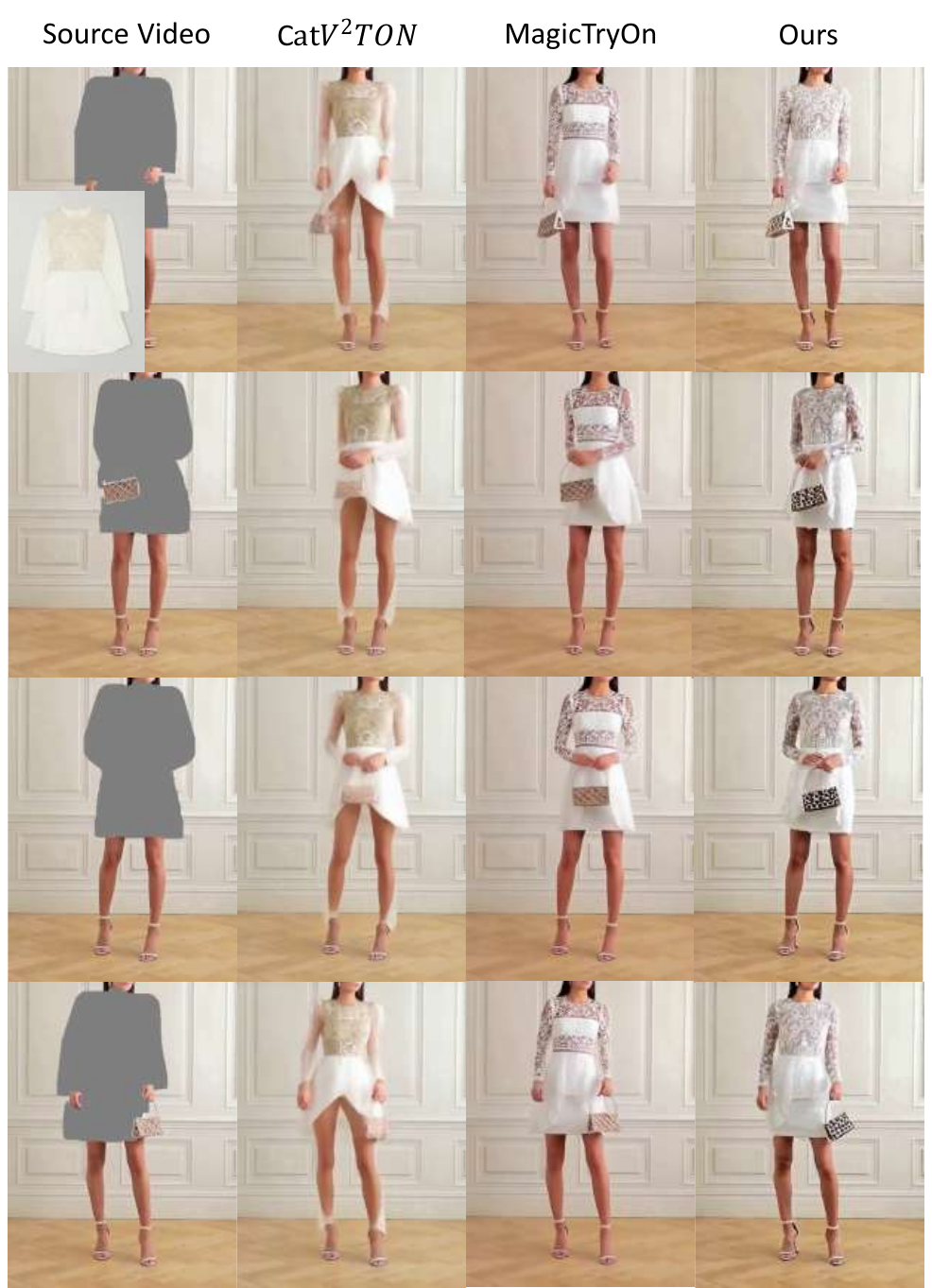}
\caption{
\textbf{Visualization Comparison.} OIE outperforms other models in garment detail control and pose consistency.
}
\vspace{-5pt}
\label{fig:visual}
\end{figure}

The effectiveness of OIE stems from the strong correlation between its first-frame editing strategy and initial output quality, along with the maturity of image-based virtual try-on and the temporal priors offered by large-scale DiT. 
Table~\ref{tab:comparision} demonstrates that OIE maintains leading performance on quantitative metrics in video virtual try-on. Consistent with Table~\ref{tab:comparision}, the visual examples in Figure~\ref{fig:visual} exhibit pronounced preservation of structural details in the ankle and upper garment regions, further substantiating OIE’s superior capability in maintaining both garment fidelity and background consistency. 

\subsection{Ablation Study}
\begin{table}[h]
\centering
\vspace{-5pt}
\caption{\textbf{Ablation study of each component on the ViViD dataset.}}
\label{tab:ablation}
\vspace{-5pt}
\small
\resizebox{0.8\linewidth}{!}{
\begin{tabular}{c|c|c|c}
\toprule
\textbf{Variations} & \textbf{VFID$^p$ $\downarrow$} & \textbf{SSIM $\uparrow$} & \textbf{LPIPS $\downarrow$} \\
\midrule
w/o pose, agnostic & 74.837 & 0.6709 & 0.6623 \\
w/o pose & 70.4268 & 0.6607 & 0.6547 \\
w/o agnostic & 10.3869 & 0.7970 & 0.0883 \\
OIE & 9.3983 & 0.8466 & 0.0774 \\
\bottomrule
\end{tabular}
}
\vspace{-12pt}
\end{table}
In Table.~\ref{tab:ablation}, we conduct ablation studies on agnostic and pose components to analyze the impact of pose guidance and mask guidance on generation. The experimental results show that pose control significantly enhances the ability to control human poses, ensuring accurate handling of complex body movements in videos. Mask guidance notably improves overall fidelity and similarity. The performance gain brought by pose control to some extent demonstrates that our approach can effectively leverage the temporal content priors from large-scale DiT models while reducing the computational burden of transfer training.

\section{Conclusion}
\label{sec:conclusion}
In this paper, we propose OIE, a novel video virtual try-on architecture designed to address key challenges faced by DiT-based approaches in clothing transfer tasks—namely, the enormous model parameters and computational cost.
Extensive experimental evaluations demonstrate that OIE achieves an excellent trade-off between performance and computational efficiency.
We believe that the proposed method can serve as valuable references for future research in this domin.

\vfill\pagebreak

\begin{small}
\bibliographystyle{IEEEbib}
\bibliography{main}
\end{small}

\end{document}